\documentclass[letterpaper, 10 pt, conference]{ieeeconf}  

\IEEEoverridecommandlockouts                              

\usepackage{graphics} 
\usepackage{epsfig} 
\usepackage{mathptmx} 
\usepackage{times} 
\usepackage{amsmath} 
\usepackage{amssymb}  
\usepackage{algorithm}
\usepackage{siunitx}
\usepackage[dvipsnames]{xcolor}
\usepackage{algpseudocode}
\usepackage{array}
\usepackage{cite}
\usepackage{subcaption}
\usepackage{tablefootnote}
\usepackage{hyperref}
\DeclareMathOperator{\sign}{sign}
\newcommand{\compressParag}{\looseness=-1}
\title{\LARGE \bf
Model Predictive Contouring Control for Vehicle Obstacle Avoidance at the Limit of Handling Using Torque Vectoring*
}

\author{Alberto Bertipaglia$^{1}$, Davide Tavernini$^{2}$, Umberto Montanaro$^{2}$,\\ Mohsen Alirezaei$^{3}$,  Riender Happee$^{1}$, Aldo Sorniotti$^{4}$ and Barys Shyrokau$^{1}$
\thanks{*The Dutch Science Foundation NWO-TTW supports the research within the EVOLVE project (nr. 18484). European Union’s Horizon 2020 research and innovation programme under the Marie Skłodowska-Curie actions, under grant agreement Nr. 872907.}
\thanks{$^{1}$Alberto Bertipaglia, Riender Happee and Barys Shyrokau are with the Department of Cognitive Robotics, Delft University of Technology, 2628 CD Delft, The Netherlands
        {\tt\small \{A.Bertipaglia, R.Happee, B.Shyrokau\}@tudelft.nl}}%
\thanks{$^{2}$Davide Tavernini and Umberto Montanaro are with the Centre for Automotive Engineering, University of Surrey, GU2 7XH Guildford, U.K.
        {\tt\small \{d.tavernini, u.montanaro\}@surrey.ac.uk}}%
\thanks{$^{3}$Mohsen Alirezaei is with the Department of Mechanical Engineering, University of Eindhoven, 5612 AZ Eindhoven, The Netherlands
        {\tt\small m.alirezaei@tue.nl}}%
\thanks{$^{4}$Aldo Sorniotti is with the Departement of Mechanical and Aerospace Engineering, Politecnico di Torino, 10129 Torino, Italy 
        {\tt\small aldo.sorniotti@polito.it}}%
}

\begin{document}

\maketitle
\thispagestyle{empty}
\pagestyle{empty}

\begin{abstract}
This paper presents an original approach to vehicle obstacle avoidance. It involves the development of a nonlinear Model Predictive Contouring Control, which uses torque vectoring to stabilise and drive the vehicle in evasive manoeuvres at the limit of handling. The proposed algorithm combines motion planning, path tracking and vehicle stability objectives, prioritising collision avoidance in emergencies. The controller's prediction model is a nonlinear double-track vehicle model based on an extended Fiala tyre to capture the nonlinear coupled longitudinal and lateral dynamics. The controller computes the optimal steering angle and the longitudinal forces per each of the four wheels to minimise tracking error in safe situations and maximise the vehicle-to-obstacle distance in emergencies. Thanks to the optimisation of the longitudinal tyre forces, the proposed controller can produce an extra yaw moment, increasing the vehicle's lateral agility to avoid obstacles while keeping the vehicle stable. The optimal forces are constrained in the tyre friction circle not to exceed the tyres and vehicle capabilities. In a high-fidelity simulation environment, we demonstrate the benefits of torque vectoring, showing that our proposed approach is capable of successfully avoiding obstacles and keeping the vehicle stable while driving a double-lane change manoeuvre, in comparison to baselines lacking torque vectoring or collision avoidance prioritisation.
\end{abstract}

%
\section{Introduction}
\begin{figure}[!t]
    \centering
    \includegraphics[width=.8\columnwidth]{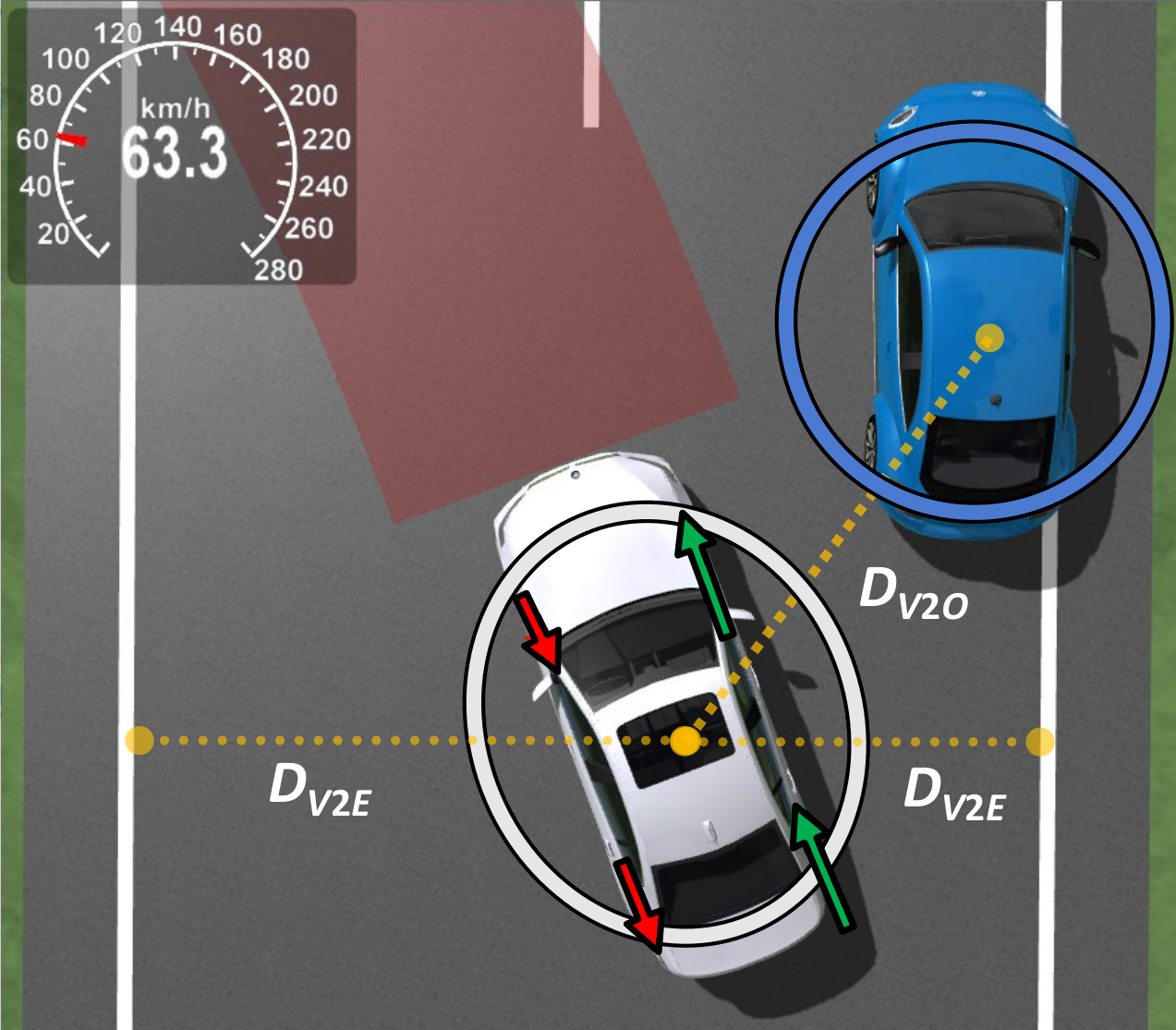}
    \caption{Instant of the vehicle controlled by the proposed MPCC during an evasive manoeuvre employing the torque vectoring capabilities.}
    \label{fig:Framework} 
\end{figure}

The ability to avoid obstacles at the limit of handling is a crucial feature to enhance the safety of automated driving vehicles. However, the highly nonlinear behaviour of the tyres, especially when the longitudinal and lateral force dynamics are coupled, makes the problem particularly challenging \cite{chowdhri2021integrated, Subosits2021Impacts}. A common approach involves developing collision avoidance controllers that optimise the steering angle and the total longitudinal brake force. However, purely braking and steering commands may not act fast enough to prevent a collision in an emergency manoeuvre \cite{Hajiloo2021Integr}. Thus, we focus on developing a collision avoidance controller that extends the classical braking and steering control inputs with torque vectoring capabilities to enhance the vehicle's lateral agility (Fig. \ref{fig:Framework}). \compressParag

Recently, a Model Predictive Contouring Control (MPCC) based on a nonlinear single-track vehicle model has been proposed for vehicle collision avoidance at the limit of handling \cite{bertipaglia2023model}. The controller integrates the motion planning, path tracking, and vehicle stability tasks into a single cost function, prioritising vehicle collision avoidance during an emergency. The MPCC describes the vehicle's kinematics using a Cartesian reference frame allowing a perfect measurement of the vehicle-to-obstacle (V2O) distance, which would be overestimated with a Frenet reference frame. Furthermore, it avoids the extra optimisation required to compute the travelled vehicle distance with respect to the reference line \cite{bertipaglia2023model, liniger2015optimization}. The controller performance is superior from a safety point of view to controllers based on hierarchical architecture \cite{Falcone2008Hiera, lenssen2023combined, Gao2014Robust} and even to an integrated architecture based on the Frenet reference system \cite{brown2019coordinating}. Despite its performance being evaluated in a high-fidelity simulation environment, the MPCC controller has never been extended or compared with a controller that utilises torque vectoring capabilities. However, new control techniques utilising torque vectoring have become particularly attractive with the commercialisation of new electric powertrain configurations, especially those based on multiple in-wheel electric motors \cite{Degel2023Scalable, de2014comparison}.

For instance, a solution is represented by a three-layer control framework: a Nonlinear Model Predictive Control (NMPC) for path tracking, a stability controller to compute the reference yaw rate, and an optimal tyre force allocation algorithm for torque vectoring \cite{ren2019integrated}. The prediction model is a single-track model with a linear tyre model. Considering the high model mismatch due to the simplistic tyre model, the stability controller layer computes a desired steady-state yaw rate to enforce vehicle stability. The torque vectoring layer allocates the optimal tyre force to each wheel, given the pre-computed desired yaw rate and longitudinal force. Simulation results demonstrate improved lateral stability and reduced path tracking error. However, the limited accuracy of a linear tyre model in emergency manoeuvres strongly reduces the performance. Furthermore, splitting the path tracking layer from reference yaw rate computation reduces the benefits of torque vectoring. For instance, an integrated path tracking and torque vectoring controller based on a Linear Quadratic Regulator achieves a higher entry speed and vehicle agility in a double lane change than a split path tracking and torque vectoring controller \cite{Chatzikomis2018Comparison}.

For this reason, a Model Predictive Control (MPC), which incorporates steering and differential braking for collision avoidance, is proposed \cite{Hajiloo2021Integr}. The framework splits the longitudinal from the lateral dynamics, and an MPC based on a linearised brush tyre model computes the desired lateral tyre force and the additional differential braking moment. However, the accuracy of a linearised tyre model strongly decreases at the limits of handling and ignores the longitudinal and lateral dynamics coupling. Furthermore, the prediction model is based on a single-track vehicle model. Thus, the maximum differential braking yaw moment is limited by a portion of the tyre force capacity to protect the vehicle's lateral force capacity from the unmodelled dynamics, making the controller more conservative.

Another approach is based on a single-track vehicle model, controlling the steering angle, the braking force distribution and an additional yaw moment \cite{stanoEnhanced2023}. The controller is capable of driving the vehicle at the limit of handling, but the limited prediction horizon \SI{1}{s} and the cost function do not allow any trajectory replanning in case of emergency.

This paper proposes an MPCC based on a nonlinear double-track vehicle with an extended Fiala tyre model for collision avoidance at the limit of handling. The proposed controller, recently proposed for collision avoidance \cite{bertipaglia2023model}, is extended to use as input not only the steering angle but also four independent longitudinal forces, generating the torque vectoring capabilities. Considering the added complexity of computing the additional yaw moment, the prediction model is refined to fully capture the lateral load transfer and the coupling between the longitudinal and lateral dynamics. The Fiala tyre model is improved to capture the variation of the cornering stiffness depending on the vertical and longitudinal force, and its saturation region is adapted to have a positive or negative gradient. The performance of the proposed controller is assessed in a high-fidelity simulation environment by performing a double lane change manoeuvre with two obstacles.

The contributions of this paper are twofold. The first is the development of the first MPCC controller extended with torque vectoring capabilities that can safely avoid a vehicle-to-obstacle collision in a double-lane change manoeuvre. At the same time, the current state-of-the-art \cite{bertipaglia2023model, brown2019coordinating} would lead to a crash. The enhanced vehicle's responsiveness due to the torque vectoring pushes the vehicle away from the obstacle, keeping it stable. The second contribution is developing and using the extended Fiala tyre model, which captures the cornering stiffness variation to longitudinal and vertical force and the gradient of the saturation tyre working region. Thus, it improves the prediction model accuracy, maximising the potential benefits of torque vectoring.

This paper is organised as follows: Section \ref{Sec:Prediction} presents the prediction model, Section \ref{Sec:MPCC} describes the proposed algorithm, Section \ref{Sec:Simulation} shows the experimental setup, and results are summarised in Section \ref{Sec:Results}. Section \ref{Sec:Conclusion} concludes the key findings and future works.
\section{Prediction Model}
\label{Sec:Prediction}
This section describes the prediction model implemented in the proposed MPCC controller. At first, the nonlinear double-track vehicle model is presented. Second, the extended Fiala tyre model with the proposed improvements is presented.

\subsection{Double-Track Vehicle Model}
\begin{figure}[!t]
    \centering
    \includegraphics[width=1\columnwidth, keepaspectratio]{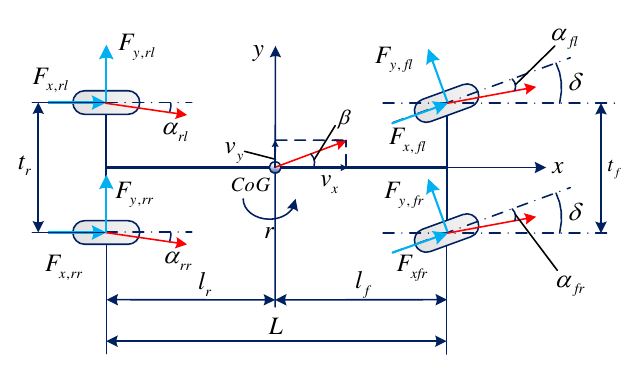}
    \caption{Double-track vehicle model.}
    \label{fig:double_track}
\end{figure}
\begin{table}[!t]
    \caption{Vehicle parameters description.}
    \label{tab:VehParameters}
    \centering
    \begin{tabular}{ccc}
        \hline
        \textbf{Parameters} & \textbf{Symbol}  & \textbf{Value} \\
        \hline
         Vehicle mass & $m$ & \SI{1997}{kg}\\
         Vehicle inertia around the z-axis & $I_{zz}$ & \SI{3198}{kg m^2}\\
         Distance between the front axle to CoG & $l_f$ & \SI{1.430}{m}\\
         Distance between the rear axle to CoG & $l_r$ & \SI{1.455}{m}\\
         Front axle track width & $t_f$ & \SI{1.540}{m}\\
         Rear axle track width & $t_r$ & \SI{1.576}{m}\\
         Air density & $\rho$ & \SI{1.204}{kg/m^3}\\
         Drag coefficient & $C_{d1}$ & \SI{0.25}{ }\\
         Rolling resistance & $C_{d0}$ & \SI{45}{N}\\
         Vehicle frontal area & $A_{f}$ & \SI{2.4}{m^2}\\
         \hline
    \end{tabular}
\end{table}
The nonlinear double-track vehicle model (Fig.\ref{fig:double_track}), is implemented in the proposed MPCC. It is chosen over the single-track vehicle model \cite{brown2019coordinating, Hajiloo2021Integr, kasinathan2015optimal} due to its capacity to capture the lateral weight transfer and its higher accuracy at the vehicle limit of handling \cite{Subosits2021Impacts}. On the other hand, the roll and pitch dynamics are ignored. The prediction model is described by twelve states: $\left(x = [X, Y, \psi, v_x, v_y, r, \theta, \delta, F_{x,\,fl}, F_{x,\,fr}, F_{x,\,rl}, F_{x,\,rr}]\right)$. The Cartesian reference system describes the vehicle's position and orientation: longitudinal position $\left(X \right)$, lateral position $\left(Y \right)$, and the heading angle $\left(\psi \right)$ of the vehicle CoG relative to an inertial frame. The longitudinal and lateral velocities at the CoG are represented, respectively, by $\left(v_x\right)$ and $\left(v_y\right)$, and the yaw rate by $\left(r\right)$. The MPCC needs to know the vehicle travelled distance $\left(\theta\right)$, which the cost function uses to compute the vehicle position relative to the reference line, so $\theta$ is added as an extra state \cite{bertipaglia2023model}.
The road-wheel angle $\left(\delta\right)$ and the longitudinal force at the front left $\left(F_{x,\,fl}\right)$, front right $\left(F_{x,\,fr}\right)$, rear left $\left(F_{x,\,rl}\right)$, and rear right $\left(F_{x,\,rr}\right)$ wheels correspond to the integral of the control inputs. The implemented state derivatives  are computed as follows:
\begin{equation}
    \begin{cases}
        \dot{X} = v_x \cos \left(\psi\right) - v_y \sin \left(\psi\right)\\
        \dot{Y} = v_x \sin \left(\psi\right) + v_y \cos \left(\psi\right)\\
        \dot{\psi} = r\\
        \dot{v}_x = \frac{1}{m} \Big(\left(F_{x,\,fl} + F_{x,\,fr}\right) \cos\left(\delta\right) - \left(F_{y,\,fl} + F_{y,\,fr}\right) \sin\left(\delta\right) + \\
        \indent + F_{x,\,rl} + F_{x,\,rr} - F_{resist}\Big) + r v_y\\
        \dot{v}_y = \frac{1}{m} \Big(\left(F_{x,\,fl} + F_{x,\,fr}\right) \sin\left(\delta\right) + \left(F_{y,\,fl} + F_{y,\,fr}\right) \cos\left(\delta\right) +\\
        \indent + F_{y,\,rl} + F_{y,\,rr} \Big) -  r v_x\\
        \dot{r} = \frac{1}{I_{zz}}\Big(\left(F_{y,\,fl} + F_{y,\,fr}\right) \cos\left(\delta\right)l_f - \left(F_{y,\,rl} + F_{y,\,rr}\right) l_r + \\
         \indent\left(F_{x,\,fl} + F_{x,\,fr}\right) \sin\left(\delta\right) l_f +\frac{t_f \left(F_{x,\,fr} - F_{x,\,fl}\right) \sin\left(\delta\right)}{2} + \\
         \indent + \frac{t_f \left(F_{y,\,fl} - F_{x,\,fr}\right) \sin\left(\delta\right)}{2} + \frac{t_r \left(F_{y,\,rr} - F_{x,\,rl}\right)}{2} \Big)\\
        \dot{\theta} = \sqrt{v_x^2 + v_y^2}
    \end{cases}
    \label{eq:Double}
\end{equation}
Where $F_{x,\,ij}$ and $F_{y,\,ij}$ are the longitudinal and lateral tyre forces, $i$ stands for front $\left(f\right)$ or rear $\left(r\right)$, and $j$ stands for left $\left(l\right)$ or right $\left(r\right)$. All other vehicle parameters are reported in Table. \ref{tab:VehParameters}. Moreover, $F_{resist}$ is the aerodynamic drag and the rolling resistance computed according to the following equation:
\begin{equation}
    F_{resist} = \frac{1}{2} \rho A_{f} C_{d1} v_x^2 + C_{d0}
    \label{eq:resist}
\end{equation}
The vehicle model inputs $\left(u_v = \left[ \dot{\delta}, \dot{F}_{y,\,fl}, \dot{F}_{y,\,fr}, \dot{F}_{y,\,rl}, \dot{F}_{y,\,rr}\right] \right)$ are the rates of the previously mentioned road wheel angle and the rates of longitudinal forces applied to each wheel. The input rates are integrated into the prediction model before being applied to the vehicle, so it is possible to apply constraints and make them smooth.

The vehicle and obstacles are represented as circles so the MPCC controller can constantly monitor the vehicle-to-obstacle (V2O) distance. Similar approach is implemented for the vehicle-to-edge of the road (V2E) distance. Their Euclidean distance is computed as follows:
\begin{equation}
    D_{V2O} = \sqrt{\left( X - X_{obs}\right)^2 + \left( Y - Y_{obs}\right)^2} - r_{obs} - r_{veh}
    \label{eq:dista}
\end{equation}
where $X,\, Y$ and $X_{obs},\, Y_{obs}$ are, respectively, the longitudinal and lateral position of the vehicle and obstacle centre, and $r_{veh}$ and $r_{obs}$ are the radii of the vehicle and obstacle circles. The proposed MPCC controller aims to keep $D_{V2O}$ above a user-defined safety distance. The vehicle and the obstacles will collide if $D_{V2O}$ is lower than zero.

\subsection{Extended Fiala Tyre Model}
The lateral tyre forces for each wheel of the double-track vehicle model are captured by an extended Fiala tyre model. The classic Fiala tyre model is modified to capture the variation of cornering stiffness depending on the longitudinal and vertical force \cite{Subosits2021Impacts}, and the saturation region is adapted to include a negative gradient. The latter allows the prediction model not to overestimate the maximum lateral force when the tyre works with a high lateral slip angle, e.g. driving at the limit of handling or drifting. The extended Fiala tyre model is defined as follows:
\begin{equation}
    \begin{split}
        &F_y\left(\alpha, F_x, F_z \right) =\\
        &=\begin{cases}
            -C_{ym}\left(F_x, F_z\right) \tan\alpha + \frac{C_{ym}^2\left(F_x, F_z\right)\tan\alpha\tan|\alpha|}{3 F_{y,\,max}} + \\
            - \frac{C_{ym}^3\left(F_x, F_z\right)\tan\alpha^3}{27 F_{y,\,max}^2}, \;\;\;\;\;\;\;\;\;\;\;\;\;\;\;\;\;\;\;\;\;\;\;\;\;|\alpha|\leq\alpha_{thr}\\\
            \frac{2 C_{ym}\left(F_x, F_z\right) \left( \zeta - 1 \right) \tan\alpha}{3} - \frac{C_{ym}^2\left(F_x, F_z\right) \left( \zeta - 1 \right) \tan\alpha |\tan\alpha|}{9 F_{y,max}} + \\
            -F_{y, max} \zeta \sign(\alpha), \;\;\;\;\;\;\;\;\;\;\;\;\;\;\;\;\;\;\;\;|\alpha|>\alpha_{thr}\
        \end{cases}
    \end{split}
    \label{eq:extFiala}
\end{equation}
where $\alpha$ is the tyre slip angle, $C_y$ is the tyre cornering stiffness, which is a function of the vertical ($F_z$) and longitudinal ($F_x$) tyre force, $F_{y,\,max}$ is the maximum lateral tyre force, $\alpha_{thr}$ is the tyre slip threshold corresponding to the peak of the tyre lateral force, and $\zeta$ is a parameter defined between 0 and 2 which characterises the gradient of the saturation region. When $\alpha\leq\alpha_{thr}$, the extended Fiala model is formulated as the classic Fiala tyre model \cite{Weber2023Modeling}, while the saturated region ($\alpha>\alpha_{thr}$) is modified to have a gradient that better captures the maximum lateral force reduction with large slip angles. At the same time, the proposed model still keeps the advantages of the classical Fiala tyre, so it is fully continuous and differentiable when $\alpha=\alpha_{thr}$. A gradient different from zero in the saturation region helps numerical optimisation algorithms based on the gradient calculation to avoid derivative vanishing and to optimise the steering angle when the tyre works in the saturation region \cite{Laurense2022Long}. Furthermore, the proposed solution has a positive gradient when $\zeta\in\left[1, 2\right]$ and a negative one when $\zeta\in\left[0, 1\right]$. 

\begin{figure}[!t]
    \centering
    \begin{subfigure}{0.92\linewidth}
        \centering
        \includegraphics[width=1\linewidth, keepaspectratio]{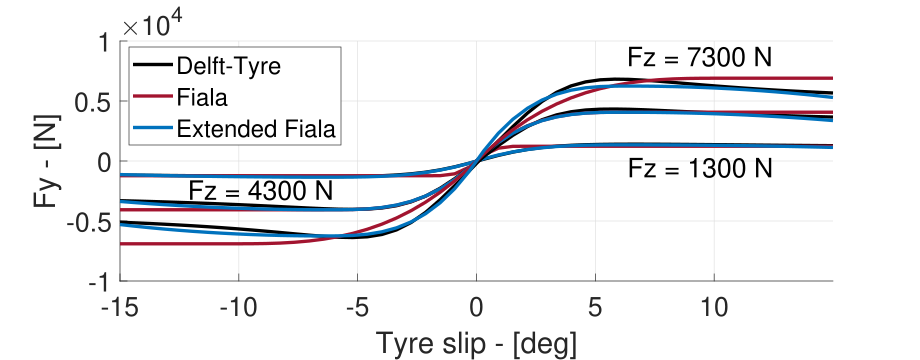}
        \caption{}
        \label{fig:FialaFz}
    \end{subfigure}
    \par\vspace{0cm}
    \begin{subfigure}{0.91\linewidth}
        \centering
        \includegraphics[width=1\linewidth, keepaspectratio]{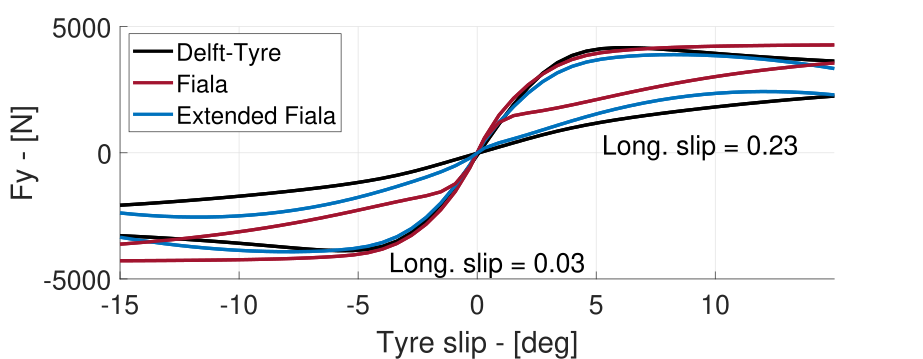}
        \caption{}
        \label{fig:FialaFx}
    \end{subfigure}
    \caption{Fig. \ref{fig:FialaFz} compares the high-fidelity Delft, the classic Fiala and the extended Fiala tyre model with different normal loads. Fig. \ref{fig:FialaFx} compares the previously mentioned models with the lateral and longitudinal force coupling.}
    \label{fig:FialaDelft}
\end{figure}

\begin{figure}[!t]
    \centering
    \begin{subfigure}{0.45\columnwidth}
        \centering
        \includegraphics[width=1\columnwidth]{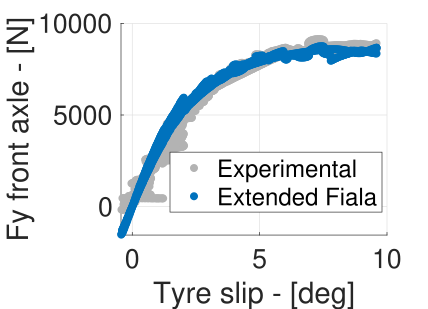}
        \caption{}
        \label{fig:FyFront}
    \end{subfigure}
    \begin{subfigure}{0.45\columnwidth}
        \centering
        \includegraphics[width=1\columnwidth]{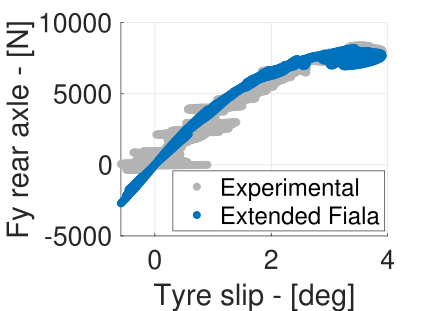}
        \caption{}
        \label{fig:FyRear}
    \end{subfigure}
    \caption{Fig. \ref{fig:FyFront} and \ref{fig:FyRear} show the extended Fiala tyre model experimental validation for the front and rear axle.}
    \label{fig:fialaExper}
\end{figure}

To further reduce the tyre model mismatch, the $C_y$ is not considered constant, but it is firstly adapted depending on the vertical force \cite{Subosits2021Impacts} as follows:
\begin{equation}
    C_y\left(F_z\right) = c_1 F_{z0} \sin\left(2\;\text{atan}\left(\frac{F_z}{c_2 F_{z0}}\right)\right)
\end{equation}
where $c_1$ and $c_2$ are tunable parameters, and $F_{z0}$ is the nominal vertical load. Second, the previously computed $C_y$ is further modified to capture its dependency from $F_x$ as follows:
\begin{equation}
    \begin{split}
        C_{ym}\left(F_x, F_z\right) =& \frac{1}{2}\left(\mu F_z  - F_x \right) +\\
        & + \left( 1 - \left( \frac{|F_x|}{\mu F_z} \right)^{c_3} \right)^{-c_3} \left( C_y\left(F_z\right) - \frac{1}{2} \mu F_z \right)
    \end{split}
\end{equation}
where $c_3$ is a user-defined parameter, and $\mu$ is the friction coefficient. The $C_{ym}\left(F_x, F_z\right)$ is used to compute the tyre slip threshold ($\alpha_{thr}$) as follows:
\begin{equation}
    \alpha_{thr} = \frac{3 F_{y,\,max}}{C_{ym}\left(F_x, F_z\right)}
\end{equation}
The maximum lateral tyre force ($F_{y,\,max}$) is limited by the tyre friction circle, defined as follows:
\begin{equation}
    F_{y,\,max} = \sqrt{\left(\mu F_z\right)^2 - F_x^2}
\end{equation}
All the parameters used in the extended Fiala tyre model are reported in Table \ref{tab:TyreParameters}, and their value is obtained by performing a nonlinear optimisation \cite{Bertipaglia2022Two} between the lateral tyre forces of the proposed extended Fiala tyre model and a high-fidelity Delft-Tyre model 6.2. Fig. \ref{fig:FialaDelft} shows how the proposed Fiala tyre model can capture the effect of the normal load on the tyre cornering stiffness and how the model mismatch between the Delft tyre model and the proposed one is reduced not only in the linear region but also around the peak lateral force area. Particularly relevant is the saturated region, which is overestimated by the Fiala tyre model with a constant saturation region \cite{Subosits2021Impacts, Weber2023Modeling, brown2019coordinating}. A model mismatch in the tyre's large slip angle working area is particularly detrimental for obstacle avoidance controllers at the limit of handling. Fig. \ref{fig:FialaDelft} shows that the cornering stiffness of the proposed Fiala captures the longitudinal and lateral force coupling, which is a significant phenomenon for the scope of this work. The reason is that the proposed MPCC uses torque vectoring capabilities, which implies using a longitudinal force coupled with a lateral one.
\begin{table}[!t]
    \caption{Tyre parameters description.}
    \label{tab:TyreParameters}
    \centering
    \begin{tabular}{ccc}
        \hline
        \textbf{Parameters} & \textbf{Symbol}  & \textbf{Value} \\
        \hline
         Lateral cornering stiffness effect & $c_1$ & \SI{49.3}{}\\
         Lateral cornering stiffness peak effect & $c_2$ & \SI{3.5}{}\\
         Long. and Lat. tyre force coupling effect & $c_3$ & \SI{4.1}{}\\
         Nominal vertical tyre force (as in .tir property file) & $F_{z0}$ & \SI{4300}{N}\\
         Tyre friction coefficient & $\mu$ & \SI{0.95}{}\\
         \hline
    \end{tabular}
\end{table}
Furthermore, the optimised tyre model is experimentally validated by performing a quasi-steady-state circular driving test. Fig. \ref{fig:fialaExper} shows how the extended Fiala tyre model perfectly captures the linear and nonlinear tyre working regions.\compressParag
\section{Model Predictive Contouring Control Using Torque Vectoring}
\label{Sec:MPCC}
This section explains how the cost function and the constraints of the proposed MPCC are formulated. Subsection \ref{Cost} focuses on describing the MPCC cost function and how it is designed to prioritise obstacle avoidance over path tracking in case of emergency. Subsection \ref{Constraints} explains how the MPCC constraints are defined to improve safety, taking into account the vehicle actuators' limitations, and avoid redundant torque vectoring utilisation.

\subsection{Cost Function with Obstacle Avoidance Prioritisation}
\label{Cost}
The proposed MPCC is based on iterative optimisation of a nonlinear cost function ($J = J_{track} + J_{inp} + J_{obs}$) \cite{bertipaglia2023model}, which is responsible for ensuring path tracking ($J_{track}$), keeping the physical feasibility of the control inputs ($J_{inp}$) and obstacle avoidance prioritisation ($J_{obs}$) in case of emergency.

The path tracking properties in the cost function are defined as follows:
\begin{equation}
    J_{track} = \sum_{i=1}^{N} \left( q_{e_{Con}} e_{Con, i}^2 + q_{e_{Lag}} e_{Lag, i}^2 + q_{e_{Vel}} e_{Vel}^2 \right) 
\end{equation}
where $N$ is the length of the prediction horizon, $e_{Con}$ is the contouring error, $e_{Lag}$ is the lag error, $e_{Vel}$ is the velocity error and $q_*$ are the weights of the respective quadratic errors. The path is followed by minimising the contouring $\left(e_{Con}\right)$ and the lag error $\left(e_{Lag}\right)$ \cite{liniger2015optimization, bertipaglia2023model}. $e_{Con}$ represents the vehicle position projection onto the desired trajectory, depending on the vehicle's travelled distance related to the reference line $\left(\theta_s\right)$. However, contrary to MPC or NMPC controllers \cite{chowdhri2021integrated} based on a Frenet reference system, the $\left(\theta_s\right)$ is unavailable for an MPCC based on a Cartesian reference frame. Thus, $\left(\theta_s\right)$ is approximated by the vehicle total travelled distance ($\theta$), and the approximation meaning is ensured by the lag error minimisation, defined as the norm between the two distances. Mathematically, the linearised $e_{Con}$ and $e_{Lag}$ are defined as follows:
\begin{equation}
    \begin{split}
        &\bar{e}_{Con} = \sin \left( \Psi_t\left(\theta\right)\right) \left(X - X_t\left( \theta \right) \right) - \cos \left( \Psi_t\left(\theta\right)\right) \left(Y - Y_t\left( \theta \right) \right)\\
        &\bar{e}_{Lag} = - \cos \left( \Psi_t\left(\theta\right)\right) \left(X - X_t\left( \theta \right) \right) - \sin \left( \Psi_t\left(\theta\right)\right) \left(Y - Y_t\left( \theta \right) \right)
    \end{split}
\end{equation}
$X_t$, $Y_t$, and $\Psi_t$ are the desired longitudinal and lateral positions and heading angle. Despite the added non-linearities and complexities of $\bar{e}_{Con}$ and $\bar{e}_{Lag}$, they allow an approximation of the Frenet reference frame with a Cartesian reference frame, which means that V2O distance is never overestimated \cite{bertipaglia2023model}. Thus, the MPCC based on a Cartesian frame is more pre-emptive than an NMPC based on a Frenet reference frame in prioritising collision avoidance over path tracking \cite{bertipaglia2023model}. Thus, the vehicle is more prone to stay inside a safe and stable working area even when it needs to avoid a collision at the limit of handling. Furthermore, the $\bar{e}_{Lag}$ is equal to zero only when the vehicle follows the reference trajectory perfectly. As soon as the desired velocity is unfeasible for the planned trajectory, the MPCC will modify the desired velocity to minimise the $\bar{e}_{Lag}$. 
The reference velocity is tracked by minimising the quadratic error between the vehicle velocity ($v_x$) and the desired one ($v_{des}$).
For what concerns the weights, they are firstly empirically tuned to reduce the path tracking error and the vehicle sideslip angle peaks \cite{bertipaglia2022model, bertipaglia2023unscented}. Second, they are fine-tuned using a two-stage Bayesian optimisation \cite{Bertipaglia2022Two}. It is important to highlight that the $q_{e_{Vel}}$ is tuned to have a low weight in magnitude because the controller must allow the vehicle to slow down in case of obstacle avoidance prioritisation \cite{bertipaglia2023model, brown2019coordinating}.

The minimisation of the control inputs is ensured in the cost function as follows:
\begin{equation}
    \begin{split}
        J_{inp} =& \sum_{i=1}^{N} \bigl( q_{\dot{\delta}} \dot{\delta_i}^2 + q_{\dot{F}_{x,\,FL}} \dot{F}_{x,\,FL, \,i}^2 + q_{\dot{F}_{x,\,FR}} \dot{F}_{x,\,FR, \,i}^2 +\\
        &+ q_{\dot{F}_{x,\,RL}} \dot{F}_{x,\,RL, \,i}^2 + q_{\dot{F}_{x,\,RR}} \dot{F}_{x,\,RR, \,i}^2 \bigl)
    \end{split}
\end{equation}
where $\dot{\delta}$ is the steering angle rate, and $\dot{F}_{x,\,FL}$, $\dot{F}_{x,\,FR}$, $\dot{F}_{x,\,RR}$, and $\dot{F}_{x,\,RR}$ are the rate of the longitudinal forces applied to each of the vehicle's four wheels. These cost terms are added in order to make the control inputs smooth.

The obstacle avoidance prioritisation is defined as follows:
\begin{equation}
    J_{obs} = \sum_{i=1}^{N} \left( \sum_{j=1}^{N_{obs}}\left( q_{e_{V2O}} e_{V2O,\,j,\, i}^2 \right) + \sum_{j=1}^{N_{edg}}\left( q_{e_{V2E}} e_{V2E,\,j,\, i}^2 \right)\right)
    \label{eq:J_obs}
\end{equation}
where $N_{obs}$ and $N_{edg}$ are the number of obstacles and road edges, $e_{V2O}$ and $e_{V2E}$ are the difference between the V2O and V2E distances and the user defined safety distances between the obstacles ($D_{Sft, O}$) and the road edges ($D_{Sft, E}$). When the vehicle is at a safe distance from obstacles or road edges, it does not interfere with the path tracking properties of the MPCC. On the other hand, the $e_{V2O}$ and $e_{V2E}$ errors allow the MPCC controller to dynamically perform a short trajectory replanning when the vehicle passes close to the obstacles. The obstacle avoidance prioritisation is due to the dynamically varying weights associated with $e_{V2O}$ and $e_{V2E}$ \cite{bertipaglia2023model}. The weights, here reported only $q_{V2O}$ for compactness, vary as follows: 
\begin{equation}
    q_{V2O} = 
    \begin{cases}
        P_{k}, & \text{if } D_{V2O} < 0 \\
        P_{k}\; e^{-\frac{2 D_{V2O}^2}{D_{Sft, O}^2 }}, & \text{elseif}\;\; 0 \leq D_{V2O} \leq D_{Sft, O} \\
        0, & \text{otherwise}
    \end{cases} \\
\end{equation}
where $P_{k}$ denotes the upper limit of the achievable value for $q_{V2O}$. The magnitude of $q_{V2O}$ increases with a Gaussian-shaped curve with the decrease of the V2O distance, and it is zero when V2O is above $D_{Sft, O}$.

\subsection{Constraints}
\label{Constraints}
The constraints are designed to accommodate actuator limitations, vehicle stability, and path tracking and to avoid redundant torque vectoring utilisation. The actuators' limitations are applied to $\delta$, $F_{x,\,FL}$, $F_{x,\,FR}$, $F_{x,\,RR}$, and $F_{x,\,RR}$ and their respective rates. The values implemented are reported in Table \ref{tab:Constraints}.

The vehicle stability is enforced using the tyre friction circle as follows \cite{bertipaglia2023model, brown2019coordinating}:
\begin{equation}
    F_{x,\,ij} = S_f \mu F_{z,\,ij}
\end{equation}
where $S_f$ is a safety factor that limits the applicable longitudinal force considering the tyre road friction coefficient uncertainty ($\mu$), and the subscripts $ij$ represent the front-rear axle and left-right side.

The following inequality forces the vehicle to stay inside the road boundaries:
\begin{equation}
    \biggl\lVert 
    \begin{bmatrix}
        X\\
        Y
    \end{bmatrix}
    - 
    \begin{bmatrix}
        X_{cen}\\
        Y_{cen}
    \end{bmatrix}
    \biggr\rVert^2 \leq \left(\frac{W_{t}}{2}\right)^2 
\end{equation}
where $X_{cen}$ and $Y_{cen}$ are the longitudinal and lateral locations of the track's centre, and $W_{t}$ is the road width \cite{liniger2015optimization}.\compressParag

The MPCC is constrained not to use redundant torque vectoring while driving in a straight to avoid excessive tyre wear and energy consumption as follows:
\begin{equation}
    \begin{split}
        & |F_{x,\,FL} - F_{x,\,FR}| \leq |F_{z,\,FL} - F_{z,\,FR}| {T}_{s} \\
        & |F_{x,\,RL} - F_{x,\,RR}| \leq |F_{z,\,RL} - F_{z,\,RR}| {T}_{s} \\  
    \end{split}
\end{equation}
where ${T}_{s}$ is a user defined parameter which works as a safety coefficient, allowing a difference in the longitudinal forces higher than the normal load difference between the right and left sides of the vehicle. These constraints push the controller not to use torque vectoring when the car is driving straight, saving energy and tyre wear.
\begin{table}[!t]
    \caption{Upper and lower constraints \cite{bertipaglia2023model,chowdhri2021integrated}.}
    \label{tab:Constraints}
    \centering
    \begin{tabular}{cccc}
        \hline
        \textbf{Symbol}  & \textbf{Boundaries} & \textbf{Symbol}  & \textbf{Boundaries}\\
        \hline
         $\dot{\delta}$ & \SI{\pm90}{deg/s} & $\delta$ & \SI{\pm18}{deg}\\
         $\dot{F_{x,\,FL}}$ & \SI{\pm7200}{N/s} & $F_{x,\,FL}$ & \SI{\pm3600}{N}\\
         $\dot{F_{x,\,FR}}$ & \SI{\pm7200}{N/s} & $F_{x,\,FL}$ & \SI{\pm3600}{N}\\
         $\dot{F_{x,\,RL}}$ & \SI{\pm7200}{N/s} & $F_{x,\,FL}$ & \SI{\pm3600}{N}\\
         $\dot{F_{x,\,RR}}$ & \SI{\pm7200}{N/s} & $F_{x,\,FL}$ & \SI{\pm3600}{N}\\
         \hline
    \end{tabular}
\end{table}  
\section{Simulation Setup and Experimental Validation}
\label{Sec:Simulation}
This section is split into two subsections. Subsection \ref{Exp1} describes how the proposed MPCC is assessed, and subsection \ref{Exp2} explains how the high-fidelity and prediction models are validated using experimental data.
\begin{figure*}[!t]
    \centering
    \subcaptionbox*{(a)}[.33\linewidth]{%
    \includegraphics[width=\linewidth]{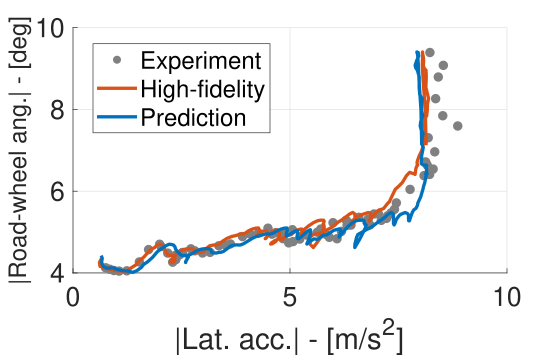}%
    }%
    \hfill
    \subcaptionbox*{(b)}[.33\linewidth]{%
    \includegraphics[width=\linewidth]{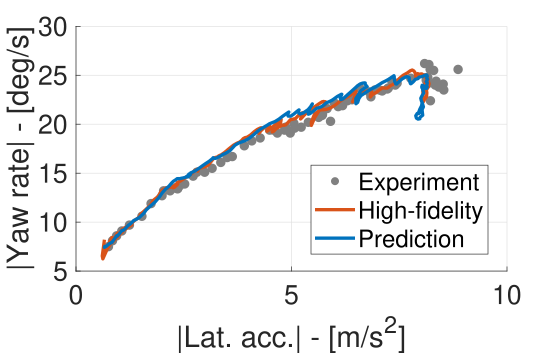}%
    }
    \hfill
    \subcaptionbox*{(c)}[.33\linewidth]{%
    \includegraphics[width=\linewidth]{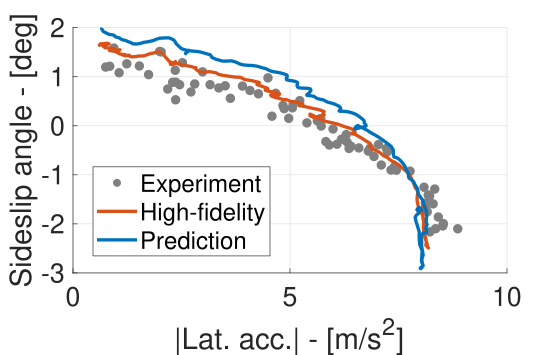}%
    }%
    \caption{Experimental model validation of the vehicle characteristics for a skidpad manoeuvre with \SI{40}{m} radius: (a) understeer gradient, (b) yaw rate and (c) sideslip angle.}
    \label{fig:steady}
\end{figure*}
\begin{figure*}[!t]
    \centering
    \subcaptionbox*{(a)}[.33\linewidth]{%
    \includegraphics[width=\linewidth]{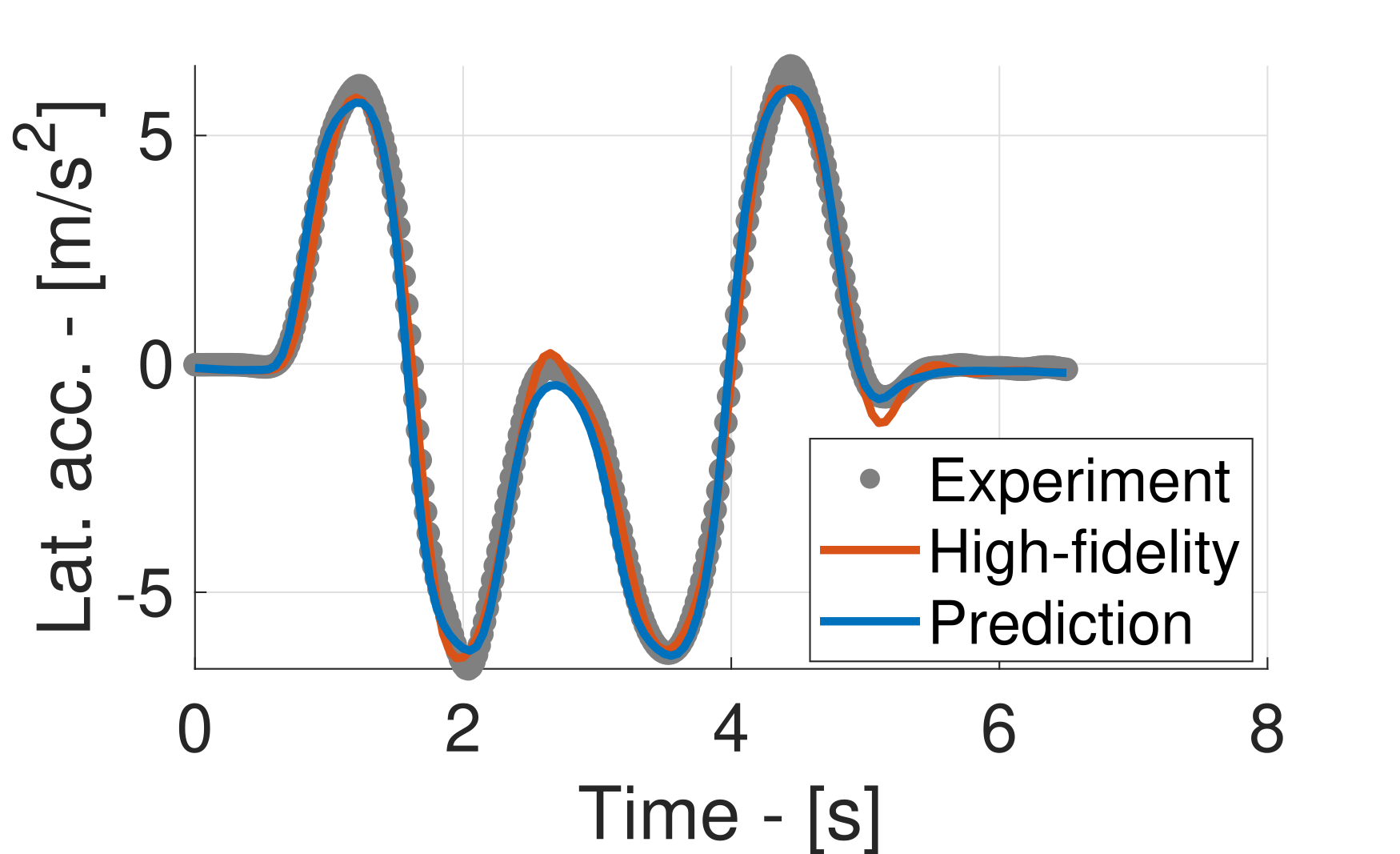}%
    }%
    \hfill
    \subcaptionbox*{(b)}[.33\linewidth]{%
    \includegraphics[width=\linewidth]{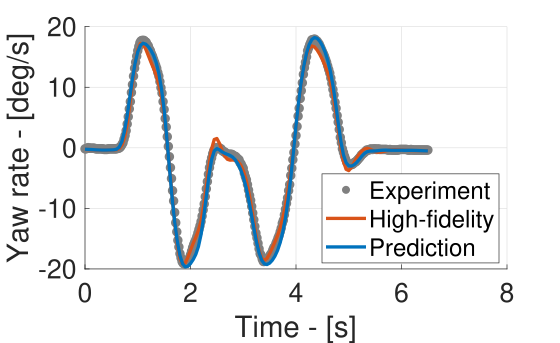}%
    }
    \hfill
    \subcaptionbox*{(c)}[.33\linewidth]{%
    \includegraphics[width=\linewidth]{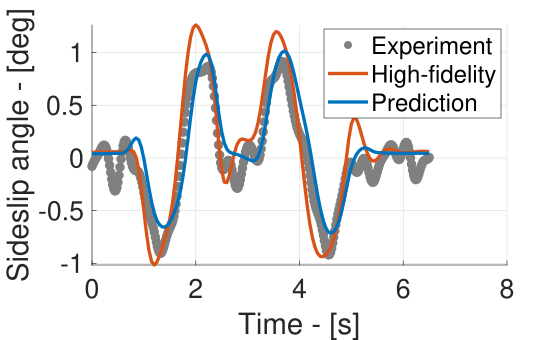}%
    }%
    \caption{Experimental model validation of the vehicle characteristics for a double lane change at \SI{70}{km/h}: (a) lateral acceleration, (b) yaw rate and (c) sideslip angle.}
    \label{fig:transient}
\end{figure*} 
\subsection{Simulation Setup}
\label{Exp1}
The proposed MPCC controller is based on a prediction horizon with \SI{30}{} steps and a sampling time of \SI{0.05}{s}. The prediction model is discretised using a Runge-Kutta 2 integration scheme, which has an optimum trade-off between accuracy and simplicity \cite{brown2019coordinating}. The nonlinear interior point solver of FORCESPro \cite{FORCESNLP} is adopted to solve the optimisation problem, and the maximum number of iterations is set to \SI{100}{}. The Broyden–Fletcher–Goldfarb–Shanno algorithm is chosen to approximate the problem Hessian, and all the other parameters are left as default.\compressParag
\begin{figure*}[htp]
    \centering
    \begin{subfigure}{2\columnwidth}
        \centering
        \includegraphics[width=1\columnwidth]{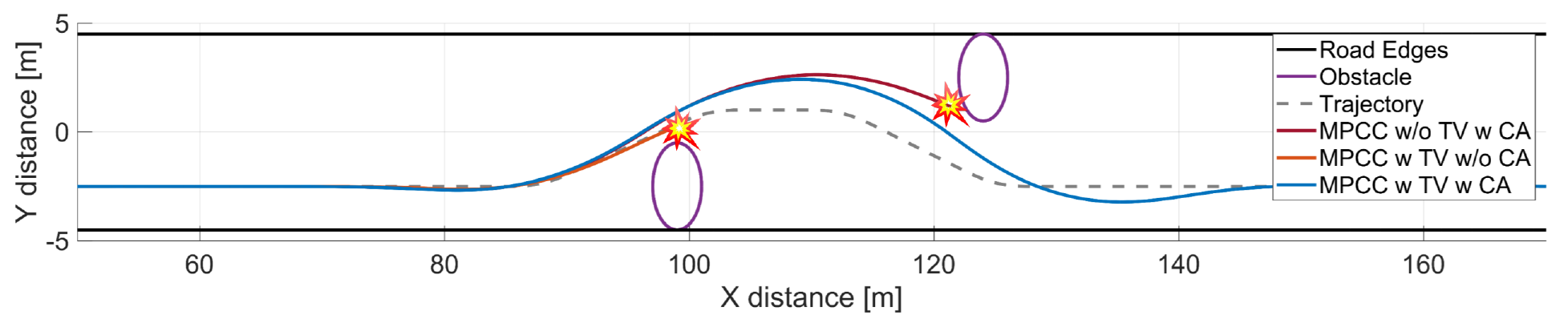}
        \caption{}
        \label{fig:tra}
    \end{subfigure}
    \hfill\\
    \begin{subfigure}{.24\linewidth}
        \centering
        \includegraphics[width=1\columnwidth]{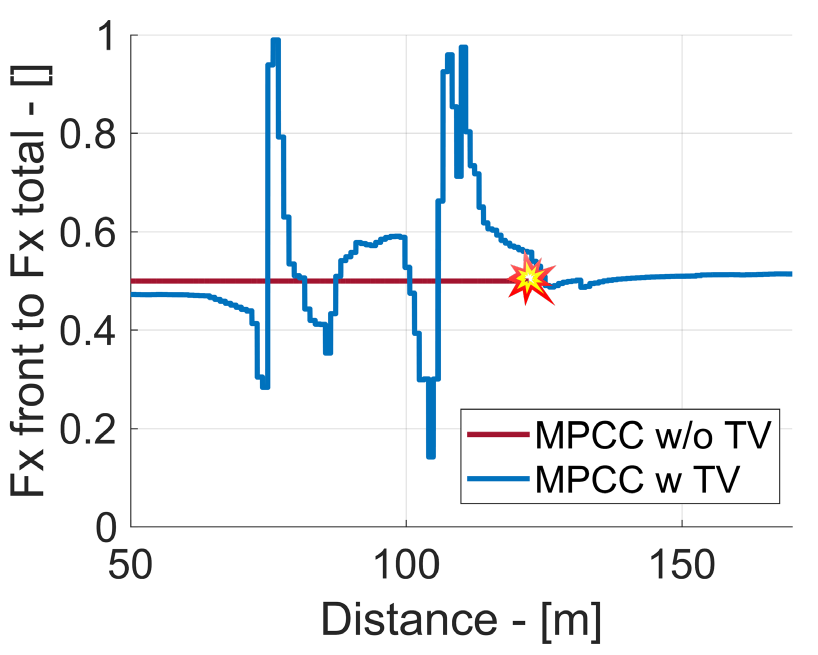}
        \caption{}
        \label{fig:ratio}
    \end{subfigure}
    \begin{subfigure}{.24\linewidth}
        \centering
        \includegraphics[width=1\columnwidth]{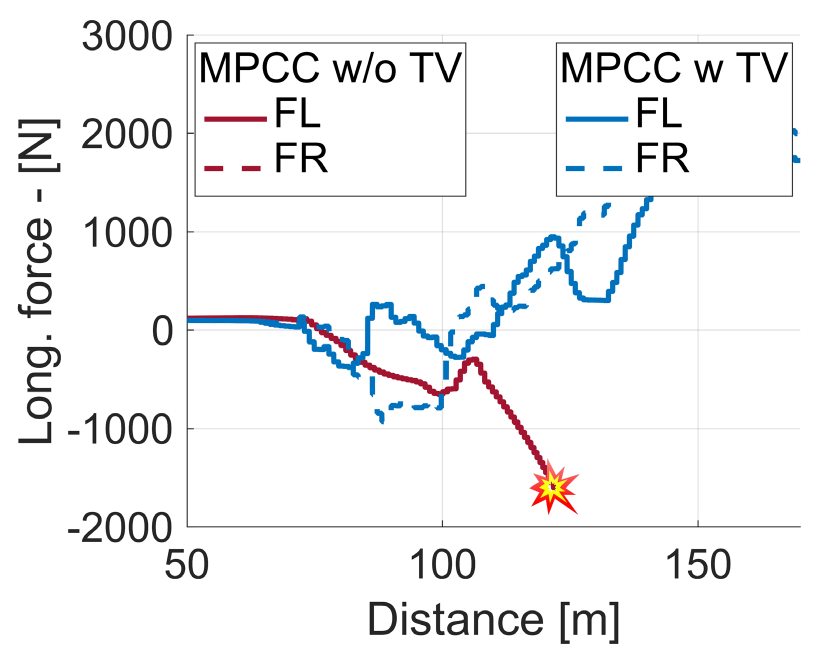}
        \caption{}
        \label{fig:FxF}
    \end{subfigure}
    \begin{subfigure}{.24\linewidth}
        \centering
        \includegraphics[width=1\columnwidth]{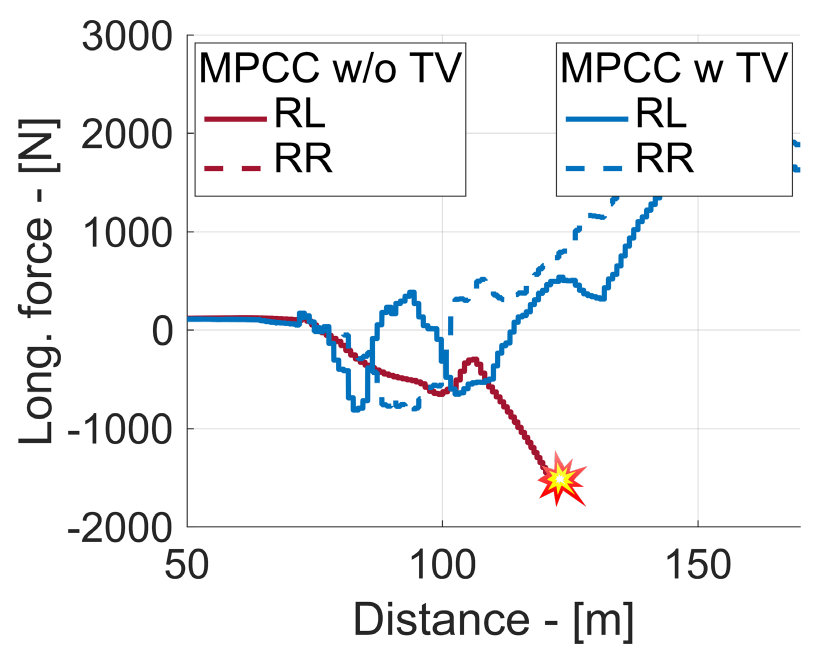}
        \caption{}
        \label{fig:FxR}
    \end{subfigure}
    \begin{subfigure}{.24\linewidth}
        \centering
        \includegraphics[width=1\columnwidth]{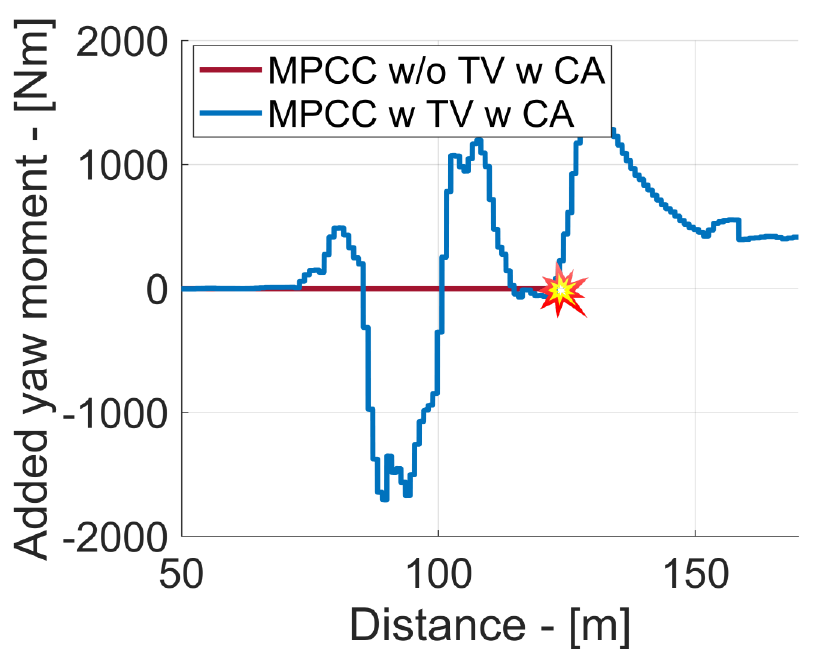}
        \caption{}
        \label{fig:moment}
    \end{subfigure}
    \hfill\\
    \begin{subfigure}{.24\linewidth}
        \centering
        \includegraphics[width=1\columnwidth]{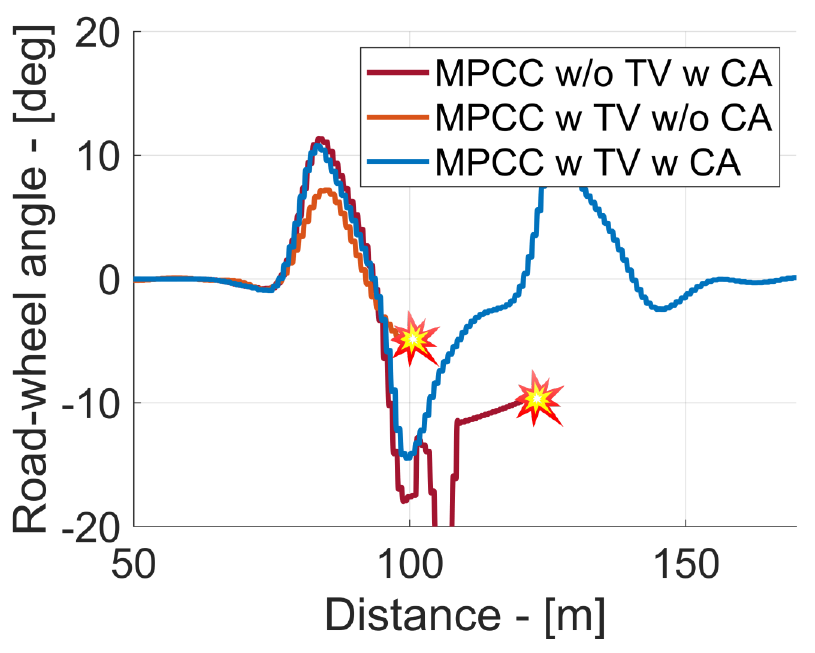}
        \caption{}
        \label{fig:delta}
    \end{subfigure}
    \begin{subfigure}{.24\linewidth}
        \centering
        \includegraphics[width=1\columnwidth]{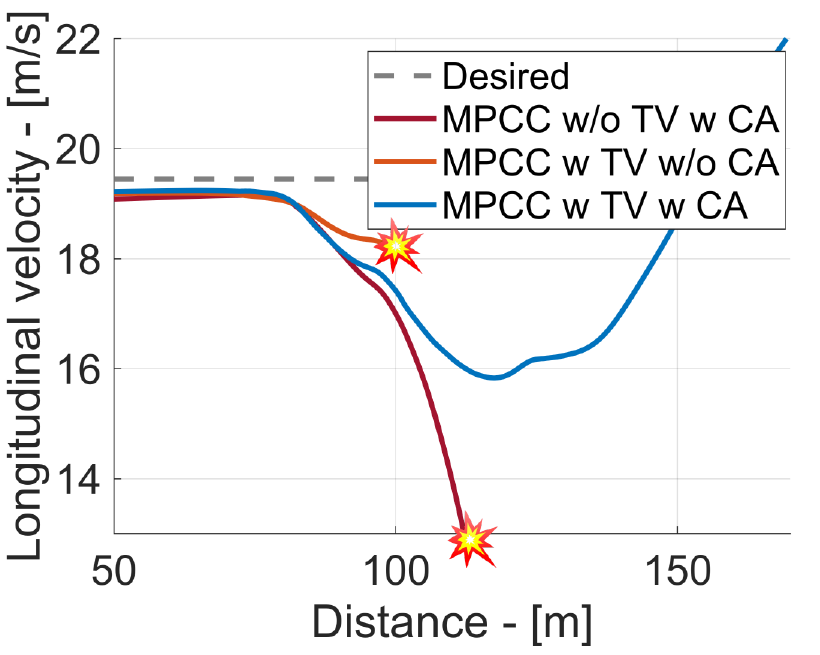}
        \caption{}
        \label{fig:Vx}
    \end{subfigure}
    \begin{subfigure}{.24\linewidth}
        \centering
        \includegraphics[width=1\columnwidth]{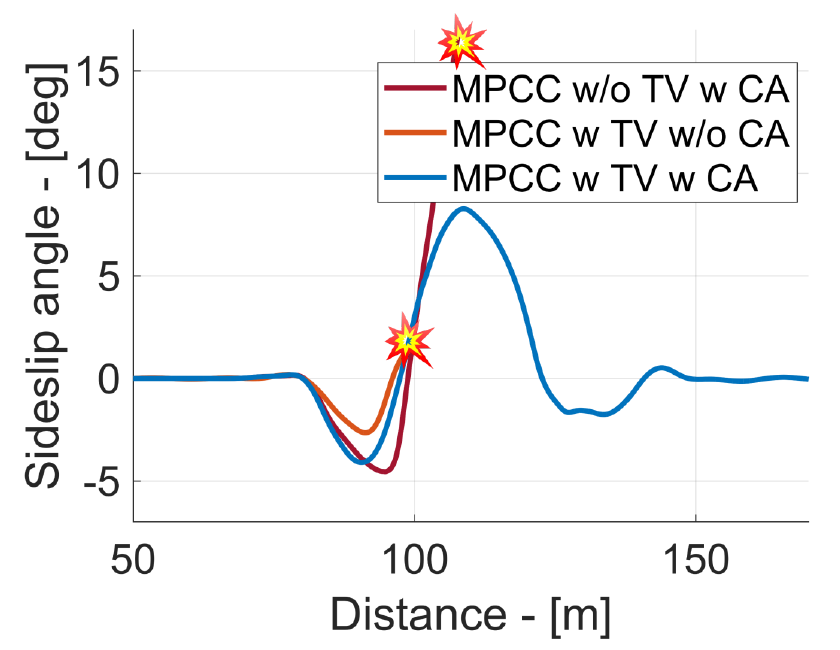}
        \caption{}
        \label{fig:sideslip}
    \end{subfigure}
    \begin{subfigure}{.24\linewidth}
        \centering
        \includegraphics[width=1\columnwidth]{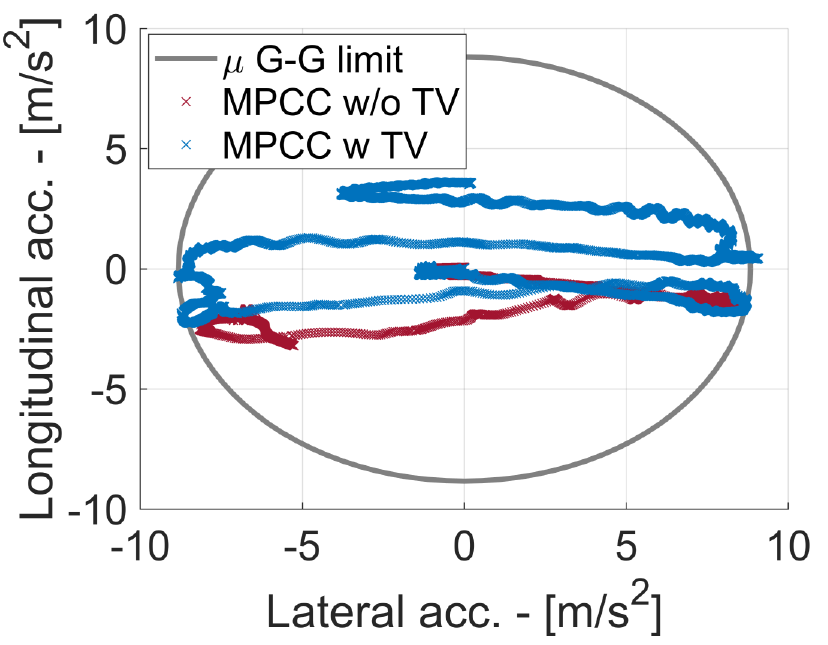}
        \caption{}
        \label{fig:GG}
    \end{subfigure} 
    \caption{States and control inputs of assessed controllers in a double lane change manoeuvre.}
    \label{fig:DoubleLane}
\end{figure*}
The proposed MPCC controller with torque vectoring and collision avoidance properties (MPCC\textsubscript{wTV, wCA}) is compared with two baselines: the same MPCC controller without the torque vectoring capabilities (MPCC\textsubscript{w/oTV, wCA}) to evaluate its benefits in evasive manoeuvres and an MPCC with torque vectoring but without the collision avoidance (MPCC\textsubscript{wTV, w/oCA}) properties of eq. \ref{eq:J_obs} to evaluate the efficacy of the obstacle avoidance prioritisation. All the controllers are evaluated on a double lane change manoeuvre with two obstacles. A coarse trajectory function of the road distance and the road curvature is provided to the MPCC, simulating the output of a behavioural planner \cite{bertipaglia2023model}. The desired velocity is assumed constant along the manoeuvre. The controller must track the reference trajectory and perform an online trajectory replanning when the vehicle passes close to one of the two obstacles. The scenario contains two obstacles to evaluate that the short trajectory replanning around the first obstacle does not interfere with the vehicle's capacity to avoid the subsequent obstacle. As a vehicle plant, a high-fidelity BMW Series 545i vehicle model is implemented on IPG CarMaker. The vehicle parameters are determined using experimental measurements in a proving ground; the suspension parameters are tuned using Kinematics \& Compliance test rig measurements, and the tyre dynamics are modelled using the Delft-Tyre 6.2 model. To further improve the plant fidelity, the steering dynamics are included through a second-order transfer function \cite{yu2021mpc}. The electric motors have faster dynamics than a conventional powertrain, and the first-order transfer function is tuned by measurements on a powertrain rig by the electric motor manufacturer. The electric motor has a time constant of \SI{\sim 6}{ms}, and the initial inverter reaction and torque ripple can be neglected \cite{Parra2021OnNo, vidal2022On, shao2023design}.
\subsection{Experimental Model Validation}
\label{Exp2}
The vehicle prediction and high-fidelity vehicle models are validated with experimental data collected on a proving ground. Fig. \ref{fig:steady} and Fig. \ref{fig:transient} show excellent model accuracy during a \SI{40}{m} radius skidpad and in the double lane change at \SI{70}{km/h}. In the quasi-steady state behaviour, the prediction and the high-fidelity model are accurate in both linear and nonlinear working regions. Regarding transient vehicle behaviour, both models can capture even the lateral acceleration, and the yaw rate peaks recorded with the experimental vehicle. The most significant difference is noticeable in Fig. \ref{fig:transient}c, which shows that the high-fidelity model overestimates the measured vehicle sideslip angle of a quantity lower than \SI{0.5}{deg}. However, the small discrepancy does not interfere with vehicle model validation.
\section{Results}
\label{Sec:Results}
Fig. \ref{fig:tra} shows the trajectories attained by the three different controllers. The proposed MPCC\textsubscript{wTV, wCA} is the only controller to drive through the double lane change manoeuvre without colliding with one of the two obstacles. The MPCC\textsubscript{wTV, w/oCA} is unaware of the obstacles' location, so its path tracking and vehicle stability objectives lead the vehicle to collide with the first obstacle located at \SI{99}{m}. On the other hand, the MPCC\textsubscript{w/oTV, wCA} can successfully avoid the first obstacle, replanning the initial trajectory in a very similar way to the MPCC\textsubscript{wTV, wCA}. However, it cannot avoid a collision with the second obstacle due to a loss of stability despite harsh braking from \SI{\sim105}{m} to \SI{\sim115}{m}. Vice versa, the MPCC\textsubscript{wTV, wCA} stabilises the vehicle between \SI{100}{m} and \SI{120}{m}, reducing the vehicle sideslip angle from a peak higher than \SI{15}{deg} to a peak equal to \SI{7.5}{deg} (Fig. \ref{fig:sideslip}). This is possible thanks to the extra yaw moment, up to \SI{\sim2000}{Nm}, at the vehicle CoG generated by the torque vectoring capabilities (Fig. \ref{fig:FxR}). Furthermore, it is relevant to notice that thanks to the lower sideslip angle, the vehicle controller MPCC\textsubscript{wTV, wCA} can drive through the double lane change at a higher speed than the MPCC\textsubscript{w/oTV, wCA}, without performing any harsh braking after avoiding the first obstacle. Fig. \ref{fig:Vx} shows that the minimum speed of the vehicle driven by the MPCC\textsubscript{wTV, wCA} is \SI{16}{m/s}, while the MPCC\textsubscript{w/oTV, wCA} cannot avoid the second obstacle despite reducing the speed to \SI{13}{m/s}. 
The MPCC\textsubscript{wTV, wCA} also optimises the front and rear longitudinal force repartition (Fig. \ref{fig:ratio}). The controller moves the brake repartition to the front axle during the hard braking at \SI{90}{m}. However, the front-rear ratio is restored to \SI{50}{\%} or even moved to the rear axle, when the vehicle enters the corner, and the front axle has a high road wheel angle (Fig. \ref{fig:delta}). It is relevant to notice that in both the front and rear axle (Fig. \ref{fig:FxF} and Fig. \ref{fig:FxR}), the added yaw moment generated by the torque vectoring is relevant in magnitude only during the manoeuvre and at its exit to stabilise the vehicle. On the contrary, the added inequality constraints constrain the MPCC\textsubscript{wTV, wCA} not to use torque vectoring when the vehicle is driving straight. 
Fig. \ref{fig:GG} shows that both MPCC controllers with CA capabilities reach the maximum lateral acceleration, which can be generated with the available road friction coefficient. On the contrary, the maximum braking capability is not fully exploited by any controller. This work demonstrates that the TV can be integrated into an MPCC with CA avoidance prioritisation and that the TV is essential to stabilise the vehicle while avoiding obstacles at the limit of handling. However, the strong coupling between longitudinal and lateral dynamics brings complexity to the MPCC. For this reason, it was essential to improve the accuracy of the prediction model, e.g. including an extended version of the Fiala tyre model.
\section{Conclusion}
\label{Sec:Conclusion}
This paper presents an original approach to vehicle obstacle avoidance. It is based on a nonlinear Model Predictive Contouring Control, which employs torque vectoring capabilities to stabilise the vehicle, performing evasive manoeuvres at the limit of handling. The proposed controller with torque vectoring and collision avoidance successfully avoids the two obstacles in the Double Lane Change manoeuvre. The first baseline controller without collision avoidance collides with the first obstacle due to its lack of motion replanning capability. The second baseline MPCC without torque vectoring cannot stabilise the vehicle after avoiding the first obstacle, and it collides with the second one. The proposed controller produces an extra yaw moment up to \SI{\sim2000}{Nm} in magnitude, increasing the vehicle's lateral agility to avoid obstacles and minimise the vehicle sideslip angle equal to a peak of \SI{7.5}{deg} rather than \SI{16}{deg} for the second baseline controller without torque vectoring. However, the proposed controller requires a more accurate prediction model, so an extended Fiala tyre model is developed. Future works involve the experimental validation of the proposed controller and the study of its sensitivity analysis to external disturbances, e.g. road friction coefficient and perception uncertainties.
\addtolength{\textheight}{-5cm}   


\bibliographystyle{IEEEtran}
\bibliography{IEEEabrv,references}

\end{document}